\documentclass{article}

\usepackage{tikz}
\usepackage{tikz-dependency}
\usepackage{float}
\usepackage{latexsym}
\usepackage{amsmath}

\usepackage{graphicx}
\usepackage{natbib}
\usepackage{url}
\usepackage{xcolor}
\usepackage{framed}
\usepackage{gb4e}
\usepackage{tabularx}
\usepackage{qtree}
\usepackage{pbox}
\usepackage{authblk}
\usepackage{rotating}
\usepackage{ulem}

\definecolor{shadecolor}{named}{lightgray}
\title{Converting the Suggested Upper Merged Ontology to Typed First-order Form}

\author[1]{Adam Pease}
\affil[1]{Articulate Software}
\affil[ ]{\textit {apease@articulatesoftware.com}}
\date{\vspace{-5ex}}

\begin{document}

\maketitle

\section{Introduction}

Imagine a voice-enabled household robot that can pick up objects and transfer
them from room to room. The owner might say "Pick up my red cart and put it in
the garage." but the robot might hear either "Pick up my red car and put it in
the garage."  or the original sentence.  A reasoning system might disambiguate
the utterance or correct it by reasoning that a typical car weights 1.5 tons and
that's 3000 pounds and the carrying capacity of the robot is 100 pounds and the
alternate text must be what was said.  A system must construct an answer to a
question that has never been asked before ("Can the robot carry a car?") perform
some simple computation involving unit conversions, to understand that 3000lbs
\textgreater 100lbs and possibly explain its answer if required ("Why didn't you
move my car like I asked you to?")

Theorem
proving
(or Automated Reasoning) is the process by which we can answer a question posed
to a theory in a mathematic logic.  It differs from search and information
retrieval (such as that done with Google or Apache Lucene) in that small facts
and rules are combined to synthesize an answer to a question that may never have
been asked before, as opposed to matching a query to the most similar previous
query and answer.  Theorem proving also provides an explanation (a proof) of how
the answer was derived.  In this work, we use the Suggested Upper Merged
Ontology (SUMO) \citep{np01,p11} as the theory, or collection of statements in
mathematical logic. There are many kinds of mathematical logic.  One of the most
popular for automated reasoning, providing a good compromise between the
expressiveness and efficiency of the language, is called First Order Logic (FOL)
(or predicate calculus).

General purpose first-order theorem provers historically haven't done proofs
with arithmetic, which is related to problems with Russel's
Paradox \citep{BaldwinLessmann}. The solution to the paradox, which was
discovered by Ernst Zermelo, involved separating numbers and other things into
different types. The solution in automated reasoning has now been largely
addressed with a language called Typed First-order Form
(TFF) \citep{SutcliffeTFF}\footnote{note that this is actually a family of logics
and specifically we translation to TF0 - Typed First-order Form with level 0
polymorphism} and implemented in several of the best modern provers,
including Vampire \citep{Kovacs:2013:FTP:2958031.2958033} from the University of
Manchester. Writing a translator from SUMO's native formalization into TFF
should open up many new opportunities for doing practical automated reasoning involving
numbers and arithmetic.

To have useful and non-trivial reasoning about, for example, a robot's
capabilities, or to do question answering, we need not only a language capable of
arithmetic calculation (as well as first order logical reasoning) but also a
non-trivial body of axioms that has the information about the real world needed
to form answers to such questions.  For that reason, we need to use the SUMO,
which is a comprehensive and diverse set of logical statements about the world.
At approximately 20,000 concepts and 80,000 logical statements, it is large
enough to answer interesting questions about a wide range of topics.

In earlier work, we described \citep{ps14} how to translate SUMO to the strictly
first order language of TPTP \citep{Sutcliffe:2007:TTC:2391910.2391914}.  SUMO
has an extensive type structure and all relations have type restrictions on
their arguments.  Translation to TPTP involved implementing a sorted (typed)
logic axiomatically in TPTP by altering all implications in SUMO to contain type
restrictions on any variables that appear.  Many other translation steps were
needed that are described in our earlier paper and part of the translation to
TFF but since they are not specific to the TFF translation they are not repeated
here in detail. Note also that all the strictly higher-order content in SUMO is
lost in translation to first-order, whether TPTP or TFF. Briefly however, the 
translation steps include:

\begin{itemize}
\item expanding "row variables" which allow for stating axioms without commitment
to the number of arguments a relation has, similar to Lisp's @REST construct
\item instantiating "predicate variables" with all possible values.  This is needed
for any axiom that has a variable in place of a relation.
\item expanding the arity of all variable arity relations as set of relations with
different names depending upon their fixed number of arguments
\item renaming any relations given as arguments to other relations
\end{itemize}

The interested reader is refered to the earlier paper for more details and examples.

\section{Typed First-order Form}

Like TPTP, TFF forms are valid Prolog syntax (although obviously not the same semantics)
TFF has five disjoint \textit{sorts}: integers, real numbers, rational numbers,
booleans and everything else. These are respectively called \$int, \$real,
\$rat, \$o and \$i in TFF syntax.  Each variable that is used in a logical
statement must be declared to be one of these sorts, or by default it will be
assumed to be type \$i.

TFF has built in to the language the basic arithmetic functions and arithmetic comparison
operators.  Each function and operator is \textit{polymorphic} - it is actually a set
of three different operators that can handle integers, rationals and reals.  Equality
is also defined for \$o and \$i.

TFF's creators have planned to include the ability to define subtypes (subsorts)
but this is not yet defined in specification or implemented in any prover.  An
additional issue is that since all types are disjoint, and SUMO allows multiple
inheritance, there is a mismatch between the two type systems.  So we have to
continue to implement much of SUMO's sort system axiomatically in TFF as in
TPTP, but have a special treatment of integers, rationals and reals that does
use the TFF type system, so we can use its arithmetic and comparison operators.

\section{Approach}

\begin{figure}[H]
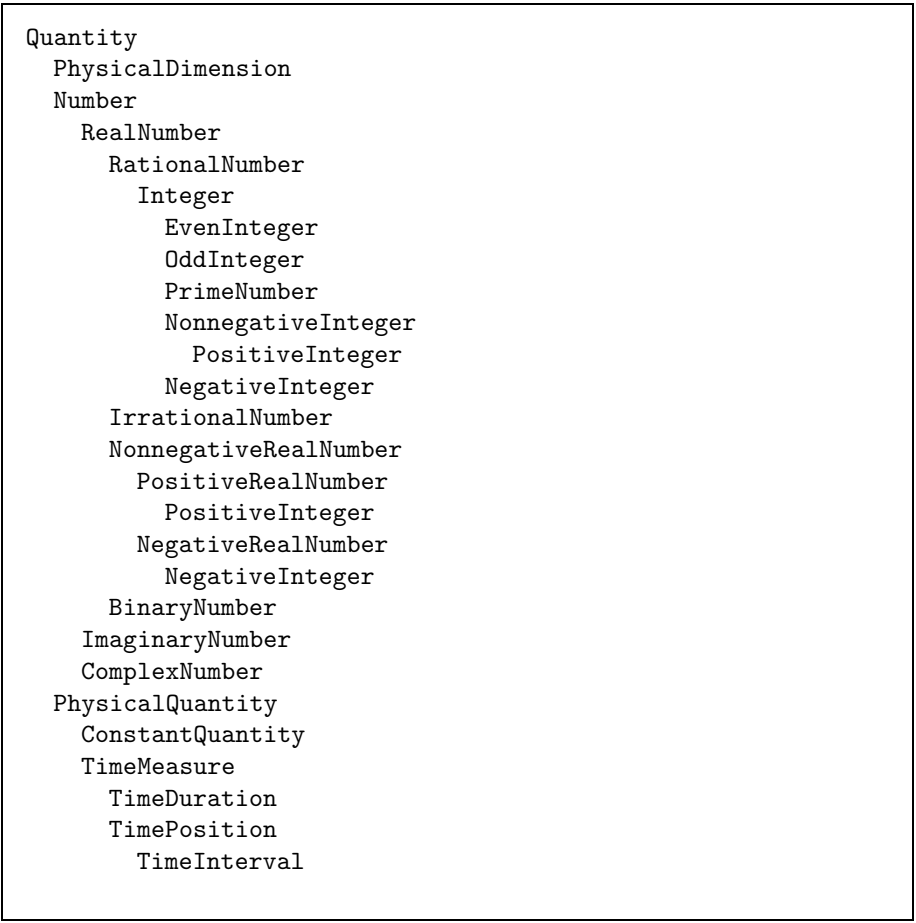

\begin{framed}
\begin{verbatim}
Quantity
  PhysicalDimension
  Number	
    RealNumber
      RationalNumber
        Integer
          EvenInteger
          OddInteger
          PrimeNumber
          NonnegativeInteger
            PositiveInteger
          NegativeInteger
      IrrationalNumber
      NonnegativeRealNumber
        PositiveRealNumber
          PositiveInteger
        NegativeRealNumber
          NegativeInteger
      BinaryNumber
    ImaginaryNumber
    ComplexNumber
  PhysicalQuantity
    ConstantQuantity
    TimeMeasure
      TimeDuration
      TimePosition
        TimeInterval
\end{verbatim}
\end{framed}
\caption{SUMO's Hierarchy of Quantity Types (somewhat simplified)}
\label{fig:QuantHier}
\end{figure}

SUMO's mathematical operators take a generic \texttt{Quantity} type, which then
has \texttt{PhysicalQuantity} (a number with units of measure) as well as
\texttt{RealNumber}(s) and \texttt{Integer}(s) as subclasses (among others see
figure \ref{fig:QuantHier}).  So we have rules like figure \ref{fig:Particulate}

\begin{figure}[H]
\begin{framed}
\begin{verbatim}
(=>
  (and
    (instance ?PM ParticulateMatter)
    (part ?Particle ?PM)
    (approximateDiameter ?Particle
      (MeasureFn ?Size Micrometer))
    (greaterThan 10 ?Size)
    (greaterThan ?Size 2.5))
  (instance ?PM CoarseParticulateMatter)))
\end{verbatim}
\end{framed}
\caption{A rule with an arithmetic comparison}
\label{fig:Particulate}
\end{figure}

where \texttt{greaterThan} is comparing integers or reals, as well as rules like
figure \ref{fig:Desert}

\begin{figure}[H]
\begin{framed}
\begin{verbatim}
(=>
  (and
    (instance ?AREA DesertClimateZone)
    (subclass ?MO Month)
    (averageTemperatureForPeriod ?AREA ?MO ?TEMP)
    (greaterThan ?TEMP
      (MeasureFn 18 CelsiusDegree)))
  (instance ?AREA SubtropicalDesertClimateZone))
\end{verbatim}
\end{framed}
\caption{A rule comparing numbers with units}
\label{fig:Desert}
\end{figure}

where \texttt{greaterThan} is comparing quantities (numbers with units) like "18
degrees".

\subsection{Type Promotion}

Since TFF doesn't have a type hierarchy, we can't have relations with numbers
and quantities (numbers with units) both allowed for their arguments. So the
approach is to separate relations that use numbers from those that don't.  

Take every relation that has a domain specification of \texttt{Quantity} or a
subclass of \texttt{Quantity}.  If the domain is a subclass of \texttt{Integer},
then promote it to \texttt{Integer} and add a constraint function to the
antecedent.  If it's a \texttt{RationalNumber} leave the type as is.  If it's
any subclass of \texttt{NonnegativeRealNumber} that's also not a subclass of
\texttt{Integer}, promote it to \texttt{RealNumber}.  All axioms containing SUMO
relations that have corresponding TFF relations will require copying with a
native TFF relation if the SUMO relation is a \texttt{Quantity} but not a
\texttt{RationalNumber}, \texttt{RealNumber} or \texttt{Integer}.

\subsection{Type Adjustment Algorithm}

Step-by-step the algorithm is

\begin{enumerate}

\item \label{alg1:vartype} Collect the types of all variables

\item \label{alg1:constrain} Constrain all types further if an equality or
inequality (an instance of SUMO's \texttt{RelationExtendedToQuantities}) has one
argument that is more specific than the stated argument type for the relation,
unless it's already a \texttt{RealNumber} (in which case even if the other
argument is an \texttt{Integer}, don't constrain it).

\item \label{alg1:promote} "promote" types that are specializations of
\texttt{Integer} or \texttt{RealNumber} as described above

\item \label{alg1:modint} if a number is an \texttt{Integer} (without a decimal
point) and is not a parameter or in a comparison statement with a defined
\texttt{Integer} type, add
".0" to it so it's a real and record its type as a \texttt{RealNumber}

\item \label{alg1:rename} Rename the \texttt{RelationExtendedToQuantities} with a
suffix if its arguments have been
constrained to those types (or their subclasses).  These will appear as the argument
number followed by one of "In", "Re" or "Ra" to stand for \texttt{Integer},
\texttt{RealNumber} or \texttt{RationalNumber} respectively.

\item \label{alg1:double} If there are any variable types for the axiom at this stage
that are below \texttt{Quantity} but not below \texttt{Number} in the SUMO
hierarchy, create two version of the axiom - one with all the original names of
the predicates and one with every SUMO predicate without a TFF equivalent having
a type specification suffix (as in the previous step) and those that do have a TFF 
equivalent converted to
that equivalent

\item \label{alg1:TFF} Translate to TFF syntax, including translating all relations
that have a corresponding native TFF relation or arithmetic operator, and have a
type specification suffix

\end{enumerate}

\subsection{Examples of Relevant Axioms}

Figure \ref{fig:Particulate} would then become figure \ref{fig:TFFParticulateArith} and 
figure \ref{fig:Desert} becomes figure \ref{fig:TFFParticulateArithSUMO}

\begin{figure}[H]
\begin{framed}
\begin{verbatim}
! [V_SIZE : $real, V_PARTICLE : $i, V_PM : $i] :
       (s__instance(V_PM, s__ParticulateMatter) &
        s__part(V_PARTICLE,V_PM) &
        s__approximateDiameter(V_PARTICLE,
            s__MeasureFn(V_SIZE,s__Micrometer)) &
        $greater(10.0,V_SIZE) &
        $greater(V_SIZE,2.5) =>
      s__instance(V_PM,s__CoarseParticulateMatter))
\end{verbatim}
\end{framed}
\caption{Axiom in TFF with native comparison operator}
\label{fig:TFFParticulateArith}
\end{figure}

\begin{figure}[H]
\begin{framed}
\begin{verbatim}
! [V_AREA : $i, V_MO : $i, V_TEMP : $i ]
        s__instance(V_AREA, DesertClimateZone) &
        s__subclass(V_MO, Month) &
        s__averageTemperatureForPeriod(V_AREA, V_MO, V_TEMP) &
        s__greaterThan(V_TEMP,
            s__MeasureFn(18,s__CelsiusDegree))) =>
    s__instance(V_AREA, s__SubtropicalDesertClimateZone))
\end{verbatim}
\end{framed}
\caption{Axiom in TFF with SUMO comparison operator}
\label{fig:TFFParticulateArithSUMO}
\end{figure}

We already have conversion axioms like figure \ref{fig:Joule}

\begin{figure}[H]
\begin{framed}
\begin{verbatim}
(=>
  (equal ?AMOUNT
    (MeasureFn ?X Joule))
  (equal ?AMOUNT
    (MeasureFn
      (MultiplicationFn 0.0002778 ?X) Watt)))
\end{verbatim}
\end{framed}
\caption{Unit conversion axiom}
\label{fig:Joule}
\end{figure}

So we need a new axiom to compare quantities for each equality and inequality, e.g.
figure \ref{fig:UnitCompare}

\begin{figure}[H]
\begin{framed}
\begin{verbatim}
(=>
  (and
    (equal ?Q1 (MeasureFn ?I1 ?U))
    (equal ?Q2 (MeasureFn ?I2 ?U))
    (greaterThan ?I1 ?I2))
  (greaterThan ?Q1 ?Q2))
\end{verbatim}
\end{framed}
\caption{Unit comparison axiom}
\label{fig:UnitCompare}
\end{figure}

but we can do this more succinctly with predicate variables as in figure \ref{fig:UnitCompareAll}

\begin{figure}[H]
\begin{framed}
\begin{verbatim}
(=>
  (and
    (instance ?REL RelationExtendedToQuantities)
    (equal ?Q1 (MeasureFn ?I1 ?U))
    (equal ?Q2 (MeasureFn ?I2 ?U))
    (?REL ?I1 ?I2))
  (?REL ?Q1 ?Q2))
\end{verbatim}
\end{framed}
\caption{Unit comparison axiom for all operators}
\label{fig:UnitCompareAll}
\end{figure}

which will create the preceding axiom plus all other relevant ones (figure
\ref{fig:AllRel}).

\begin{figure}[H]
\begin{framed}
\begin{verbatim}  
(instance equal RelationExtendedToQuantities)   
(instance greaterThan RelationExtendedToQuantities)
(instance greaterThanOrEqualTo RelationExtendedToQuantities)   
(instance lessThan RelationExtendedToQuantities)   
(instance lessThanOrEqualTo RelationExtendedToQuantities) 
\end{verbatim}
\end{framed}
\caption{All relations that take numbers and quantities}
\label{fig:AllRel}
\end{figure}

The axiom in figure \ref{fig:UnitCompareAll} plus the relation statements from
figure \ref{fig:AllRel} will result in the axiom in figure \ref{fig:UnitCompare}
(and many others) being generated by the predicate variable instantiation
process in Sigma.  Then when SUMO's \texttt{greaterThan} appears with strictly
integer or real arguments the subsequent TFF conversion step in Sigma will
convert it to \$greater.

When caling up to processing all of SUMO to TFF, rather than a subset, we need
to treat numeric type polymorphism as a "macro" - every statement that uses
TFF's built-in math operators.

\subsection{Detailed Example of the Algorithm}

Let's take the axiom in figure \ref{fig:RemainderFn} and the argument type
specification for the functions it contains as the start of the step-by-step
translation

\begin{figure}[H]
\begin{framed}
\begin{verbatim}
(=>
  (equal 
    (RemainderFn ?NUMBER1 ?NUMBER2) 
    ?NUMBER)
  (equal 
    (SignumFn ?NUMBER2) 
    (SignumFn ?NUMBER)))
	
(domain RemainderFn 1 Quantity)
(domain RemainderFn 2 Quantity)	
(range RemainderFn Quantity)

(domain SignumFn 1 RealNumber)	
(range SignumFn Integer)
\end{verbatim}
\end{framed}
\caption{Axiom containing functions and their type signatures}
\label{fig:RemainderFn}
\end{figure}

\begin{sloppypar}For Step \ref{alg1:vartype}, the current Sigma translation will know
to constrain each type to its most specific value in the axiom, which in this
case means that \texttt{?NUMBER} and \texttt{?NUMBER2}, since they are both
arguments to \texttt{SignumFn}, must be \texttt{RealNumber}, since
\texttt{SignumFn} takes one argument that must be a \texttt{RealNumber} -
\texttt{(domain SignumFn 1 RealNumber)}.  We'll have a list of
\texttt{[?NUMBER=RealNumber, ?NUMBER1=Quantity, ?NUMBER2=RealNumber]}
\end{sloppypar}

\begin{sloppypar}For Step \ref{alg1:constrain} we do have the
\texttt{RelationExtendedToQuantities} \texttt{RemainderFn} as well as
\texttt{equal} where we must constrain its argument to be the same lowest common
type, which among \texttt{[?NUMBER1=Quantity, ?NUMBER2=RealNumber]} is
\texttt{RealNumber}.  So the list of variables and their types becomes
\texttt{[?NUMBER=RealNumber, ?NUMBER1=RealNumber, ?NUMBER2=RealNumber]}.  Step
\ref{alg1:promote} doesn't apply since the variable list contains no specializations
of \texttt{Integer} or \texttt{RealNumber}.  Step \ref{alg1:modint} doesn't apply
since we have no literal numbers. Since we've modified the arguments of
\texttt{RemainderFn} we need to rename it in step \ref{alg1:rename} as
\texttt{RemainderFn\_\_0Re1Re2Re}.  Then we have to look at \texttt{equal}.  Note
that we have to keep track of a modified return type for every function that
returns some numerical type. In this case, although \texttt{RemainderFn} returns
a \texttt{Quantity}, its arguments have already constrained it to return a
\texttt{RealNumber}.  So we have to check that the arguments to \texttt{equal}
are the same and constrain them if needed.  In this case, \texttt{?NUMBER} is
already a \texttt{RealNumber} and doesn't need to be constrained further.
\end{sloppypar}

Step \ref{alg1:double} doesn't
apply since there are no variables of type \texttt{Number}.

Now we have figure \ref{fig:FunctExpand}

\begin{figure}[H]
\begin{framed}
\begin{verbatim}
(=>
  (equal 
    (RemainderFn__0Re1Re2Re ?NUMBER1 ?NUMBER2) 
    ?NUMBER)
  (equal 
    (SignumFn ?NUMBER2) 
    (SignumFn ?NUMBER)))
\end{verbatim}
\end{framed}
\caption{Expansion of function names}
\label{fig:FunctExpand}
\end{figure}

The last step is to translate into TFF syntax and convert relations and functions to
their native TFF names, when applicable.  This results in figure \ref{fig:TFFRemainderFn}.

\begin{figure}[H]
\begin{framed}
\begin{verbatim}
! [V_NUMBER : $real, V_NUMBER1 : $real, 
   V_NUMBER2 : $real] :
       ($equal(s__RemainderFn__0Re1Re2Re(V_NUMBER1,V_NUMBER2),
               V_NUMBER) =>
       ($equal(s__SignumFn(V_NUMBER2),
               s__SignumFn(V_NUMBER))))
\end{verbatim}
\end{framed}
\caption{Axiom in TFF with native comparison operator}
\label{fig:TFFRemainderFn}
\end{figure}

Let's take an example axiom that maps to some TFF
built in functions or relations.  In the case of \texttt{greaterThan}, we may
have SUMO axioms in which it is used with units, and cases where it is used
without.  The cases without units will map to TFF's built in \$greater relation,
as in figure \ref{fig:NumCompare}

\begin{figure}[H]
\begin{framed}
\begin{verbatim}
(=>
  (and
    (instance ?X ?Y)
    (subclass ?Y PureSubstance)
    (boilingPoint ?Y
      (MeasureFn ?BOIL KelvinDegree))
    (meltingPoint ?Y
      (MeasureFn ?MELT KelvinDegree))
    (measure ?X
      (MeasureFn ?TEMP KelvinDegree))
    (greaterThan ?TEMP ?MELT)
    (lessThan ?TEMP ?BOIL))
  (attribute ?X Liquid))
\end{verbatim}
\end{framed}
\caption{Another rule with numeric comparisons}
\label{fig:NumCompare}
\end{figure}

\begin{sloppypar} Step \ref{alg1:vartype} results in the following where a '+'
signifies a class rather than an instance - \texttt{[?X=Quantity,
?Y=PureSubstance+, ?BOIL=RealNumber, ?TEMP=RealNumber, ?MELT=RealNumber]}. For
Step \ref{alg1:constrain} we have \texttt{greaterThan} and \texttt{lessThan} but
all are \texttt{RealNumber} so no change is needed. Step \ref{alg1:promote}
doesn't apply. For Step \ref{alg1:modint} there are no numbers so that doesn't
apply.   Step \ref{alg1:rename} we have \texttt{greaterThan} and
\texttt{lessThan}, both have \texttt{RealNumber} as arguments and so will get
renamed to \texttt{greaterThan\_\_0Re1Re2Re} and \texttt{lessThan\_\_0Re1Re2ReReal} respectively,
as shown in Figure \ref{fig:RenameReal}. Step \ref{alg1:double} doesn't apply.
For Step \ref{alg1:TFF} we rename \texttt{greaterThan\_\_0Re1Re2Re} to \$greater and
\texttt{lessThan\_\_0Re1Re2Re} to \$less and translate to TFF syntax, resulting in
Figure \ref{fig:TFFRenameReal}. \end{sloppypar}

\begin{figure}[H]
\begin{framed}
\begin{verbatim}
(=>
  (and
    (instance ?X ?Y)
    (subclass ?Y PureSubstance)
    (boilingPoint ?Y
      (MeasureFn ?BOIL KelvinDegree))
      (meltingPoint ?Y
        (MeasureFn ?MELT KelvinDegree))
      (measure ?X
        (MeasureFn ?TEMP KelvinDegree))
      (greaterThan__0Re1Re2Re ?TEMP ?MELT)
    (lessThan__0Re1Re2Re ?TEMP ?BOIL))
  (attribute ?X Liquid))
\end{verbatim}
\end{framed}
\caption{Rule with numeric comparisons after renaming with Real}
\label{fig:RenameReal}
\end{figure}

\begin{figure}[H]
\begin{framed}
\begin{verbatim}
! [V_X : $i, V_Y : $i, V_BOIL : $real, 
   V_TEMP : $real, V_MELT : $real] :
      (s__instance(V_X,V_Y) &
       s__subclass(V_Y,s__PureSubstance) &
       s__boilingPoint(V_Y,
                       s__MeasureFn(V_BOIL,s__KelvinDegree)) &
       s__meltingPoint(V_Y,
                       s__MeasureFn(V_MELT,s__KelvinDegree)) &
       s__measure(V_X,
                  s__MeasureFn(V_TEMP,s__KelvinDegree)) &
       $greater(V_TEMP,V_MELT) &
       $less(V_TEMP,V_BOIL)) =>
      s__attribute(V_X,s__Liquid)
\end{verbatim}
\end{framed}
\caption{Rule in TFF with numeric comparisons after renaming operators}
\label{fig:TFFRenameReal}
\end{figure}

Another case is figure \ref{fig:LiquidDrop}

\begin{figure}[H]
\begin{framed}
\begin{verbatim}
(<=>
  (and
    (instance ?LD LiquidDrop)
    (approximateDiameter ?LD
      (MeasureFn ?Size Micrometer))
    (lessThan 500 ?Size))
  (instance ?LD Droplet))
\end{verbatim}
\end{framed}
\caption{Axiom with literal numbers}
\label{fig:LiquidDrop}
\end{figure}

For Step \ref{alg1:vartype} we have \texttt{[?LD=Droplet,?Size=RealNumber]}. For
step \ref{alg1:constrain}, since \texttt{?Size} is already a \texttt{RealNumber}
it doesn't need to be changed to be a \texttt{RealNumber} like '500.0' given
that they are both arguments to \texttt{lessThan}.  Step \ref{alg1:promote}
doesn't apply. For Step \ref{alg1:modint} we promote '500' to be '500.0'. In
Step \ref{alg1:rename} we change \texttt{lessThan} to \texttt{lessThan\_\_0Re1Re2Re} to
conform to its argument types.   Step \ref{alg1:double} doesn't apply.
For step \ref{alg1:TFF} we make \texttt{lessThan\_\_0Re1Re2Re} into \$less and do the
translation to TFF syntax that results in figure \ref{fig:TFFLiquidDrop}.

\begin{figure}[H]
\begin{framed}
\begin{verbatim}
! [V_Size : $real, V_LD : $i] :
  ((s__instance(V_LD,s__LiquidDrop) & 
    s__approximateDiameter(V_LD,
                       s__MeasureFn(V_Size,s__Micrometer)) &
    $less(500.0,V_Size)) =>
  s__instance(V_LD,s__Droplet)))

! [V_Size : $real, V_LD : $i] :
  (s__instance(V_LD,s__Droplet) =>
   (s__instance(V_LD,s__LiquidDrop) & 
    s__approximateDiameter(V_LD,
                       s__MeasureFn(V_Size,s__Micrometer)) &
    $less(500.0,V_Size)) out
\end{verbatim}
\end{framed}
\caption{TFF Axiom with literal numbers}
\label{fig:TFFLiquidDrop}
\end{figure}

\begin{figure}[H]
\begin{framed}
\begin{verbatim}
(=>
    (instance ?X NegativeInteger)
    (greaterThan 0 ?X))
\end{verbatim}
\end{framed}
\caption{Axiom with NegativeInteger}
\label{fig:NegInt}
\end{figure}

For Figure \ref{fig:NegInt} the transformation steps are, in order

\begin{enumerate}
\item \texttt{[?X=NegativeInteger]}
\item Doesn't apply
\item Promote \texttt{NegativeInteger} to \texttt{Integer}, so we have \texttt{[?X=Integer]}
\item Doesn't apply
\item We make \texttt{greaterThan} into \texttt{greaterThan\_\_0In1In2In}
\item Doesn't apply
\item TFF syntax translation results in figure \ref{fig:TFFNegInt}
\end{enumerate}

\begin{figure}[H]
\begin{framed}
\begin{verbatim}
! [V_X : $int] :
    (s__instance(V_X,s__NegativeInteger) =>
     $greater(0,V_X))
\end{verbatim}
\end{framed}
\caption{TFF Axiom with NegativeInteger}
\label{fig:TFFNegInt}
\end{figure}

But this doesn't work since \texttt{instance} takes \$i types and not \$int
which is the type of \texttt{V\_X}.  

\section{Modification to the Translation Algorithm}

In effect we need to suppress all axioms
involving types below \texttt{Integer} and \texttt{RealNumber}, but only
if they are the definitions of those terms!  Otherwise, we need to have the definitions
become restrictions on integers and reals.  So we build a table of definitional
constraints for all types below \texttt{Integer} and \texttt{RealNumber}.  This
will consist of figure \ref{fig:NumberDefs1} and \ref{fig:NumberDefs2}.

\begin{figure}[H]
\begin{framed}
\begin{verbatim}
(=>
    (instance ?NUMBER EvenInteger)
    (equal
        (RemainderFn ?NUMBER 2) 0))

(=>
    (instance ?NUMBER OddInteger)
    (equal
        (RemainderFn ?NUMBER 2) 1))

(=>
    (instance ?PRIME PrimeNumber)
    (forall (?NUMBER)
        (=>
            (equal
                (RemainderFn ?PRIME ?NUMBER) 0)
            (or
                (equal ?NUMBER 1)
                (equal ?NUMBER ?PRIME)))))

(=>
    (instance ?X NonnegativeInteger)
    (greaterThan ?X -1))

(=>
    (instance ?X NegativeInteger)
    (greaterThan 0 ?X))

(=>
    (instance ?X PositiveInteger)
    (greaterThan ?X 0))

(<=>
    (instance ?NUMBER PositiveRealNumber)
    (and
        (greaterThan ?NUMBER 0)
        (instance ?NUMBER RealNumber)))	

(=>
    (instance ?NUMBER PositiveRealNumber)
    (equal
        (SignumFn ?NUMBER) 1))

\end{verbatim}
\end{framed}
\caption{Number definitions}
\label{fig:NumberDefs1}
\end{figure}

\begin{figure}[H]
\begin{framed}
\begin{verbatim}
(<=>
    (instance ?NUMBER NegativeRealNumber)
    (and
        (lessThan ?NUMBER 0)
        (instance ?NUMBER RealNumber)))	

(=>
    (instance ?NUMBER NegativeRealNumber)
    (equal
        (SignumFn ?NUMBER) -1))

(<=>
    (instance ?NUMBER NonnegativeRealNumber)
    (and
        (greaterThanOrEqualTo ?NUMBER 0)
        (instance ?NUMBER RealNumber)))		
(=>
    (instance ?NUMBER NonnegativeRealNumber)
    (or
        (equal
            (SignumFn ?NUMBER) 1)
        (equal
            (SignumFn ?NUMBER) 0)))
\end{verbatim}
\end{framed}
\caption{Number definitions (continued)}
\label{fig:NumberDefs2}
\end{figure}

That leaves some types of numbers that have no axioms other than definitions and
which are not used as argument types: \texttt{IrrationalNumber},
\texttt{BinaryNumber}, \texttt{ImaginaryNumber} and \texttt{ComplexNumber}.  We
will exclude these from translation, at least for now.  Note also that there are
two subtypes that inherit from \texttt{RealNumber} and \texttt{Integer}.  They are
\texttt{PositiveInteger}, which inherits from \texttt{PositiveRealNumber} and 
\texttt{NegativeInteger}, which inherits from \texttt{NegativeRealNumber}.  We will
come back to this issue later.

We reformulate some of the axioms to be consistent, removing the bi-implications
and the type assertions when they occur on both sides.  This gives us a few revised
axioms as shown in figure \ref{fig:RevNumberDefs}.

\begin{figure}[H]
\begin{framed}
\begin{verbatim}
(=>
    (instance ?NUMBER PositiveRealNumber)
    (greaterThan ?NUMBER 0))

(=>
    (instance ?NUMBER NegativeRealNumber)
    (lessThan ?NUMBER 0))	

(=>
    (instance ?NUMBER NonnegativeRealNumber)
    (greaterThanOrEqualTo ?NUMBER 0))
\end{verbatim}
\end{framed}
\caption{Revised number definitions}
\label{fig:RevNumberDefs}
\end{figure}

We preprocess all axioms looking for antecedents that specify an instance of a
\texttt{Number} then cache the consequent in a map indexed by the type.  We also
extract the variable name that constrains the type in those antecedents.  We wind up
with the list of antecedents and variables, respectively, shown in figure \ref{fig:NumTypeConAll}

\begin{figure}[H]
\begin{framed}
\begin{verbatim}
{PositiveInteger =
   (greaterThan ?X 0), 
 OddInteger = 
   (equal (RemainderFn ?NUMBER 2) 1), 
 NegativeRealNumber = 
   (equal (SignumFn ?NUMBER) -1), 
 EvenInteger = 
   (equal (RemainderFn ?NUMBER 2) 0),
 NonnegativeInteger = 
   (greaterThan ?X -1), 
 NegativeInteger = 
   (greaterThan 0 ?X), 

 PrimeNumber = 
   (forall (?NUMBER) 
     (=> 
       (equal (RemainderFn ?PRIME ?NUMBER) 0) 
       (or 
         (equal ?NUMBER 1) 
         (equal ?NUMBER ?PRIME)))), 

 RationalNumber = 
   (exists (?INT1 ?INT2) 
     (and 
       (instance ?INT1 Integer) 
       (instance ?INT2 Integer) 
       (equal ?NUMBER (DivisionFn ?INT1 ?INT2)))), 

 NonnegativeRealNumber = 
   (or 
     (equal (SignumFn ?NUMBER) 1) 
     (equal (SignumFn ?NUMBER) 0))}

{PositiveInteger =       X, 
 OddInteger =            NUMBER, 
 NegativeRealNumber =    NUMBER, 
 EvenInteger =           NUMBER, 
 NonnegativeInteger =    X, 
 NegativeInteger =       X, 
 PrimeNumber =           PRIME, 
 RationalNumber =        NUMBER, 
 NonnegativeRealNumber = NUMBER}
\end{verbatim}
\end{framed}
\caption{All number type conditions}
\label{fig:NumTypeConAll}
\end{figure}

For
example, we would have \texttt{[PositiveInteger=(greaterThan ?NUMBER 0)]}.
The value of the index will be added as an antecedent to any axiom that has that
type for a variable.  For example, see figure \ref{fig:NumTypeCon} where the
first axiom becomes the second, and the resulting TFF version is the third.

\begin{figure}[H]
\begin{framed}
\begin{verbatim}
(=>
    (equal ?W
        (WeekFn ?N ?Y))
    (during ?W ?Y))

(=>
  (greaterThan ?N 0)
  (=>
    (equal ?W
        (WeekFn ?N ?Y))
    (during ?W ?Y)))

! [V_N : $int, V_W : $i, V_Y : $i] :
  ($greater(V_N,0) =>
    ((V_W = s__WeekFn(V_N,V_Y)) =>
       s__during(V_W,V_Y)))
\end{verbatim}
\end{framed}
\caption{Adding a number type condition}
\label{fig:NumTypeCon}
\end{figure}

We have a different substitution process for appearances of number subtypes in
rule consequents. If there is an instance statement involving a number subtype
in the consequent, then replace it with the defining condition for that sybtype 
\textit{and all its parents} from the list in
figure \ref{fig:NumTypeConAll}.  For example, see figure \ref{fig:NumTypeConseq},
in which a variable is concluded to be a \texttt{PositiveRealNumber} and must therefore
be replaced with its defining condition

\begin{figure}[H]
\begin{framed}
\begin{verbatim}
(=>
    (measure ?QUAKE
        (MeasureFn ?VALUE RichterMagnitude))
    (instance ?VALUE PositiveRealNumber))

(=>
    (measure ?QUAKE
        (MeasureFn ?VALUE RichterMagnitude))
    (greaterThan ?VALUE 0))

! [V_QUAKE : $i, V_VALUE : $real] : 
  (s__measure(V_QUAKE,
              s__MeasureFn(V_VALUE,s__RichterMagnitude)) =>
  $greater(V_VALUE,0)
\end{verbatim}
\end{framed}
\caption{Adding a number type in a consequent}
\label{fig:NumTypeConseq}
\end{figure}

\section{Type Propagation from Functions}

Our translation algorithm is unfortunately still incomplete, as described so far.
Type changes can also propagate throughout a formula because of functions.  Take
Figure \ref{fig:NestedFunctions} that has an axiom with several nested functions, which in turn
give the variables the types below it.  But the return types of the functions should also
constrain the types of the variables.

\begin{figure}[H]
\begin{framed}
\begin{verbatim}
(<=>
  (equal
    (RemainderFn ?NUMBER1 ?NUMBER2) ?NUMBER)
  (equal
    (AdditionFn
      (MultiplicationFn
        (FloorFn
          (DivisionFn ?NUMBER1 ?NUMBER2)) 
          ?NUMBER2) 
        ?NUMBER) 
      ?NUMBER1))

{?NUMBER1=[Quantity], ?NUMBER2=[Quantity], ?NUMBER=[Quantity]}
\end{verbatim}
\end{framed}
\caption{Nested Functions}
\label{fig:NestedFunctions}
\end{figure}

The return types of the functions are shown in Figure \ref{fig:FunctionsRetType}.  While only
\texttt{FloorFn} has types more restricted than \texttt{Quantity}, these will actually
propagate through to constrain the entire set of variables.

\begin{figure}[H]
\begin{framed}
\begin{verbatim}
(domain RemainderFn 1 Quantity)	
(domain RemainderFn 2 Quantity)
(range RemainderFn Quantity)

(domain AdditionFn 1 Quantity)
(domain AdditionFn 2 Quantity)

(domain MultiplicationFn 1 Quantity)
(domain MultiplicationFn 2 Quantity)

(domain FloorFn 1 RealNumber)	
(range FloorFn Integer)
	
(domain DivisionFn 1 Quantity)	
(domain DivisionFn 2 Quantity)
\end{verbatim}
\end{framed}
\caption{Function Argument and Return Types}
\label{fig:FunctionsRetType}
\end{figure}
 
\texttt{FloorFn} makes the return type of \texttt{DivisionFn} a
\texttt{RealNumber}, which then makes \texttt{?NUMBER1} and \texttt{?NUMBER2}
\texttt{RealNumber}s but since \texttt{?NUMBER2} is an \texttt{Integer}
\texttt{?NUMBER1} must also be an \texttt{Integer} and the return type of
\texttt{DivisionFn} becomes \texttt{Integer}, meaning we need to rename as
\texttt{DivisionFn\_\_0In1In2In} and then \texttt{FloorFn} must become
\texttt{FloorFn\_\_1In2}.  Since \texttt{MultiplicationFn\_\_0In1In2In} returns an
\texttt{Integer}, that constrains \texttt{?NUMBER} also to be an
\texttt{Integer}.  \texttt{RemainderFn} then also becomes
\texttt{RemainderFn\_\_0In1In2In}. The result is the variable set and TFF
translation shown in Figure \ref{fig:VarPropRes}.

\begin{figure}[H]
\begin{framed}
\begin{verbatim}
{?NUMBER1=[Integer], ?NUMBER2=[Integer], ?NUMBER=[Integer]}

! [V__NUMBER1 : $int,V__NUMBER2 : $int,V__NUMBER : $int] : 
  ((s__RemainderFn__0In1In2In(V__NUMBER1, V__NUMBER2) = 
   V__NUMBER => 
     $sum(
       $product(
         s__FloorFn__1In(
           $quotient_e(V__NUMBER1, V__NUMBER2)),
         V__NUMBER2),
       V__NUMBER) = 
       V__NUMBER1) &
  ($sum(
    $product(
      s__FloorFn__1In(
        $quotient_e(V__NUMBER1, V__NUMBER2)) ,
      V__NUMBER2),
    V__NUMBER) = 
    V__NUMBER1 => 
      s__RemainderFn__0In1In2In(V__NUMBER1, V__NUMBER2) = 
        V__NUMBER))
\end{verbatim}
\end{framed}
\caption{Variable Type Propagation Result}
\label{fig:VarPropRes}
\end{figure}

We then also need to copy all the axioms on the original term over to the new
term with the modified suffix.

Every relation that has an argument type between \texttt{Quantity} and
\texttt{Integer} will need and additional \texttt{Integer} version.  Those with
arguments between \texttt{Quantity} and \texttt{RealNumber} will need
\texttt{Integer} and \texttt{RationalNumber} versions. Those with arguments of
\texttt{Quantity}, \texttt{PhysicalDimension} or \texttt{Number} will need all
three - \texttt{Integer}, \texttt{RealNumber} and \texttt{RationalNumber} as
well as their original version.

\subsection{Sort Conflict}

Inequalities in TFF must be used with a single sort.  However, in SUMO we have the 
case of figure \ref{fig:TypeConflict}

\begin{figure}[H]
\begin{framed}
\begin{verbatim}
(=> 
  (equal 
    (CeilingFn ?NUMBER) 
    ?INT) 
  (not 
    (exists (?OTHERINT) 
      (and 
        (instance ?OTHERINT Integer) 
        (greaterThanOrEqualTo ?OTHERINT ?NUMBER) 
        (lessThan ?OTHERINT ?INT)))))
\end{verbatim}
\end{framed}
\caption{Type Conflict}
\label{fig:TypeConflict}
\end{figure}

\texttt{CeilingFn} takes a \texttt{RealNumber} and returns the smallest \texttt{Integer} 
greater than or equal to the argument.  The \texttt{greaterThanOrEqualTo} must
compare the \texttt{RealNumber} and \texttt{Integer} but that is not allowed in TFF.

We would like to be able to state the following in figure \ref{fig:TypeConflictTFF}

\begin{figure}[H]
\begin{framed}
\begin{verbatim}
(! [V__NUMBER : $real,V__INT : $int] : 
  (s__CeilingFn__0In1ReFn(V__NUMBER) = V__INT => 
  ~ ? [V__OTHERINT:$int] : 
    ($greatereq(V__OTHERINT ,V__NUMBER) & 
    $less(V__OTHERINT ,V__INT))))).
\end{verbatim}
\end{framed}
\caption{Type ConflictTFF}
\label{fig:TypeConflictTFF}
\end{figure}

but that generates an error due to the arguments to \texttt{\$greatereq} being of
different types.  A similar problem exists in SUMO's axiom for \texttt{FloorFn}.   
We can use TFF's type coercion function to solve this, as in figure \ref{fig:TypeConflictTFFcoerce}

\begin{figure}[H]
\begin{framed}
\begin{verbatim}
(! [V__NUMBER : $real,V__INT : $int] : 
  (s__CeilingFn__0In1ReFn(V__NUMBER) = V__INT => 
  ~ ? [V__OTHERINT:$int] : 
    ($greatereq(V__OTHERINT ,$to_real(V__NUMBER)) & 
    $less(V__OTHERINT ,V__INT))))).
\end{verbatim}
\end{framed}
\caption{Eliminating Type ConflictTFF with type coercion}
\label{fig:TypeConflictTFFcoerce}
\end{figure}

Alternatively, for these particular axioms we could use the built-in TFF
\$ceiling and \$floor functions, and then even eliminate these axioms entirely
since they help define the these TFF built-in functions.

Another version of this problem occurs when mixing \texttt{Quantity}(s) and
basic numeric types, as with

\begin{figure}[H]
\begin{framed}
\begin{verbatim}
(=> 
  (diameter ?CIRCLE ?LENGTH) 
  (exists (?HALF) 
    (and 
      (radius ?CIRCLE ?HALF) 
      (equal (MultiplicationFn ?HALF 2) ?LENGTH))))
\end{verbatim}
\end{framed}
\caption{Type ConflictTFForig}
\label{fig:ConflictTFForig}
\end{figure}

We'd like to multiply, for example, "two feet" by 2 and get "four feet".  This
is equivalent to the following -

\begin{figure}[H]
\begin{framed}
\begin{verbatim}
(=> 
  (diameter ?CIRCLE ?LENGTH) 
  (exists (?N__HALF ?U__HALF) 
    (and 
      (radius ?CIRCLE (MeasureFn ?N__HALF ?U__HALF)) 
      (equal 
        (MeasureFn (MultiplicationFn ?N__HALF 2) ?U__HALF) 
        ?LENGTH))))
\end{verbatim}
\end{framed}
\caption{Type Conflict TFF modification}
\label{fig:TypeConflictTFFmod}
\end{figure}

where the numeric part of the quantity with units is multiplied by the numeric
argument.

\section{Conclusion}

We hope that the extensive list of practical challenges in translating a large
and comprehensive theory to TFF will help motivate future version of the TFF
that may make theorem proving with arithmetic more straightforward.  In particular,
having a type hierarchy would greatly simplify the translation steps that are
needed.  

\begin{sloppypar}
The software for this translation is implemented in Java as part of
the Sigma Knowledge Engineering Environment, available with the Suggested
Upper Merged Ontology (SUMO) on GitHub at \url{https://github.com/ontologyportal}.
That site also has an expanded technical report providing additional translation
details that space doesn't permit including in this paper.
\end{sloppypar}

\section{Appendix: Sample Proof}

Here we show a sample proof with arithmetic of the sample problem from the
introduction.  We assert that a robot can only carry objects of less than
100 pounds (axiom f1) and then assert existence of an event where it carries a 1.5
ton object (axioms f2 and f4).  We pose this query to the Vampire theorem prover.  We see that
in step f71 it applies the unit conversion axiom stated in f8 to convert
100 pounds to 0.05 tons and that 1.5 is not less than 0.05.  
This yields a contradiction.  

\begin{verbatim}
~$ /home/apease/workspace/vampire/vampire --mode casc -t 300 
  /home/apease/.sigmakee/KBs/Robot-small.tff

% Refutation found. Thanks to Tanya!
% SZS status Unsatisfiable for Robot-small
% SZS output start Proof for Robot-small

tff(func_def_0, type, s__MeasureFn__1ReFn: ($real * $i) > $i).

tff(f1,axiom,(
  s__maximumPayloadCapacity(s__Robot1,
                            s__MeasureFn__1ReFn(100.0,
                                                s__PoundMass))),
  file('/home/apease/.sigmakee/KBs/Robot-small.tff',kb_SUMO_3179)).

tff(f2,axiom,(
  s__measure(s__MyCar,s__MeasureFn__1ReFn(1.5,s__TonMass))),
  file('/home/apease/.sigmakee/KBs/Robot-small.tff',kb_SUMO_3180)).

tff(f3,axiom,(
  ! [X0,X1,X2,X3] : ((greaterThan(X2,X3) & 
    s__maximumPayloadCapacity(X1,X3) & 
    s__measure(X0,X2)) => 
  ~? [X4] : (s__instance(X4,s__Carrying) & 
             s__patient(X4,X0) & 
             s__instrument(X4,X1)))),
  file('/home/apease/.sigmakee/KBs/Robot-small.tff',kb_SUMO_268)).

tff(f4,axiom,(
  s__instance(s__Carry1,s__Carrying) & 
  s__patient(s__Carry1,s__MyCar) & 
  s__instrument(s__Carry1,s__Robot1)),
  file('/home/apease/.sigmakee/KBs/Robot-small.tff',assert)).

tff(f8,axiom,(
  ! [X3 : $real] : s__MeasureFn__1ReFn(X3,s__PoundMass) = 
    s__MeasureFn__1ReFn($quotient(X3,2000.0),s__TonMass)),
  file('/home/apease/.sigmakee/KBs/Robot-small.tff',kb_SUMO_3175b2)).

tff(f9,axiom,(
  ! [X6,X7 : $real,X8 : $real] : ($greater(X7,X8) => 
    greaterThan(s__MeasureFn__1ReFn(X7,X6),s__MeasureFn__1ReFn(X8,X6)))),
  file('/home/apease/.sigmakee/KBs/Robot-small.tff',kb_SUMO_228b)).

tff(f10,axiom,(
  ! [X6,X7 : $real,X8 : $real] : 
    (greaterThan(s__MeasureFn__1ReFn(X7,X6),s__MeasureFn__1ReFn(X8,X6)) =>
    $greater(X7,X8))),
  file('/home/apease/.sigmakee/KBs/Robot-small.tff',kb_SUMO_228c)).

tff(f11,plain,(
  ! [X6,X7 : $real,X8 : $real] : ($less(X8,X7) => 
    greaterThan(s__MeasureFn__1ReFn(X7,X6),s__MeasureFn__1ReFn(X8,X6)))),
  inference(evaluation,[],[f9])).

tff(f12,plain,(
  ! [X6,X7 : $real,X8 : $real] : 
    (greaterThan(s__MeasureFn__1ReFn(X7,X6),s__MeasureFn__1ReFn(X8,X6)) => 
    $less(X8,X7))),
  inference(evaluation,[],[f10])).

tff(f14,plain,(
  ! [X0 : $real] : s__MeasureFn__1ReFn(X0,s__PoundMass) = 
    s__MeasureFn__1ReFn($quotient(X0,2000.0),s__TonMass)),
  inference(rectify,[],[f8])).

tff(f16,plain,(
  ! [X0,X1 : $real,X2 : $real] : ($less(X2,X1) => 
    greaterThan(s__MeasureFn__1ReFn(X1,X0),s__MeasureFn__1ReFn(X2,X0)))),
  inference(rectify,[],[f11])).

tff(f17,plain,(
  ! [X0,X1 : $real,X2 : $real] : 
    (greaterThan(s__MeasureFn__1ReFn(X1,X0),s__MeasureFn__1ReFn(X2,X0)) => 
    $less(X2,X1))),
  inference(rectify,[],[f12])).

tff(f20,plain,(
  ! [X0,X1 : $real,X2 : $real] : 
    (greaterThan(s__MeasureFn__1ReFn(X1,X0),s__MeasureFn__1ReFn(X2,X0)) |
    ~$less(X2,X1))),
  inference(ennf_transformation,[],[f16])).

tff(f21,plain,(
  ! [X0,X1 : $real,X2 : $real] : ($less(X2,X1) | 
    ~greaterThan(s__MeasureFn__1ReFn(X1,X0),s__MeasureFn__1ReFn(X2,X0)))),
  inference(ennf_transformation,[],[f17])).

tff(f22,plain,(
  ! [X0,X1,X2,X3] : (! [X4] : 
   (~s__instance(X4,s__Carrying) | 
    ~s__patient(X4,X0) | 
    ~s__instrument(X4,X1)) | 
    (~greaterThan(X2,X3) | 
     ~s__maximumPayloadCapacity(X1,X3) | 
     ~s__measure(X0,X2)))),
  inference(ennf_transformation,[],[f3])).

tff(f23,plain,(
  ! [X0,X1,X2,X3] : (! [X4] : 
    (~s__instance(X4,s__Carrying) | 
    ~s__patient(X4,X0) | 
    ~s__instrument(X4,X1)) | 
    ~greaterThan(X2,X3) | 
    ~s__maximumPayloadCapacity(X1,X3) | 
    ~s__measure(X0,X2))),
  inference(flattening,[],[f22])).

tff(f25,plain,(
  s__maximumPayloadCapacity(s__Robot1,s__MeasureFn__1ReFn(100.0,s__PoundMass))),
  inference(cnf_transformation,[],[f1])).

tff(f26,plain,(
  s__measure(s__MyCar,s__MeasureFn__1ReFn(1.5,s__TonMass))),
  inference(cnf_transformation,[],[f2])).

tff(f27,plain,(
  s__instrument(s__Carry1,s__Robot1)),
  inference(cnf_transformation,[],[f4])).

tff(f28,plain,(
  s__patient(s__Carry1,s__MyCar)),
  inference(cnf_transformation,[],[f4])).

tff(f29,plain,(
  s__instance(s__Carry1,s__Carrying)),
  inference(cnf_transformation,[],[f4])).

tff(f30,plain,(
  ( ! [X0:$real] : (s__MeasureFn__1ReFn(X0,s__PoundMass) = 
    s__MeasureFn__1ReFn($quotient(X0,2000.0),s__TonMass)) )),
  inference(cnf_transformation,[],[f14])).

tff(f32,plain,(
  ( ! [X2:$real,X0,X1:$real] : 
    (greaterThan(s__MeasureFn__1ReFn(X1,X0),s__MeasureFn__1ReFn(X2,X0)) | 
    ~$less(X2,X1)) )),
  inference(cnf_transformation,[],[f20])).

tff(f33,plain,(
  ( ! [X2:$real,X0,X1:$real] : ($less(X2,X1) | 
    ~greaterThan(s__MeasureFn__1ReFn(X1,X0),s__MeasureFn__1ReFn(X2,X0))) )),
  inference(cnf_transformation,[],[f21])).

tff(f34,plain,(
  ( ! [X4,X2,X0,X3,X1] : 
    (~s__instance(X4,s__Carrying) | 
     ~s__patient(X4,X0) | 
     ~s__instrument(X4,X1) | 
     ~greaterThan(X2,X3) | 
     ~s__maximumPayloadCapacity(X1,X3) | 
     ~s__measure(X0,X2)) )),
  inference(cnf_transformation,[],[f23])).

tff(f35,plain,(
  ( ! [X4,X0,X1] : 
    (~s__instance(X4,s__Carrying) | 
     ~s__patient(X4,X0) |
     ~s__instrument(X4,X1) | 
     sP0(X1,X0)) )),
  inference(cnf_transformation,[],[f35_D])).

tff(f35_D,plain,(
  ( ! [X0,X1] : (( ! [X4] : 
    (~s__instance(X4,s__Carrying) | 
     ~s__patient(X4,X0) | 
     ~s__instrument(X4,X1)) ) <=> 
    ~sP0(X1,X0)) )),
  introduced(general_splitting_component_introduction,[new_symbols(naming,[sP0])])).

tff(f36,plain,(
  ( ! [X2,X0,X3,X1] : 
    (~greaterThan(X2,X3) | 
     ~s__maximumPayloadCapacity(X1,X3) | 
     ~s__measure(X0,X2) | 
     ~sP0(X1,X0)) )),
  inference(general_splitting,[],[f34,f35_D])).

tff(f37,plain,(
  ( ! [X2,X0,X3] : 
    (sP1(X3,X0) | 
     ~s__measure(X0,X2) | 
     ~greaterThan(X2,X3)) )),
  inference(cnf_transformation,[],[f37_D])).

tff(f37_D,plain,(
  ( ! [X0,X3] : (( ! [X2] : 
    (~s__measure(X0,X2) | 
     ~greaterThan(X2,X3)) ) <=> 
    ~sP1(X3,X0)) )),
  introduced(general_splitting_component_introduction,[new_symbols(naming,[sP1])])).

tff(f38,plain,(
  ( ! [X0,X3,X1] : 
    (~sP0(X1,X0) | 
     ~s__maximumPayloadCapacity(X1,X3) | 
     ~sP1(X3,X0)) )),
  inference(general_splitting,[],[f36,f37_D])).

tff(f39,plain,(
  ( ! [X0,X1] : 
    (~s__patient(s__Carry1,X0) | 
     ~s__instrument(s__Carry1,X1) | 
     sP0(X1,X0)) )),
  inference(resolution,[],[f35,f29])).

tff(f40,plain,(
  ( ! [X0:$real,X1:$real] : 
    (~$less($quotient(X0,2000.0),X1) |   
     greaterThan(s__MeasureFn__1ReFn(X1,s__TonMass),
                 s__MeasureFn__1ReFn(X0,s__PoundMass))) )),
  inference(superposition,[],[f32,f30])).

tff(f44,plain,(
  ( ! [X0] : 
    (sP0(X0,s__MyCar) | 
     ~s__instrument(s__Carry1,X0)) )),
  inference(resolution,[],[f39,f28])).

tff(f45,plain,(
  ( ! [X2,X0:$real,X1:$real] : 
    (greaterThan(s__MeasureFn__1ReFn(X0,s__TonMass),
                 s__MeasureFn__1ReFn(X1,s__PoundMass)) |   
    ~greaterThan(s__MeasureFn__1ReFn(X0,X2),
                 s__MeasureFn__1ReFn($quotient(X1,2000.0),X2))) )),
  inference(resolution,[],[f40,f33])).

tff(f47,plain,(
  ( ! [X0,X1] : 
    (~sP1(X1,s__MyCar) | 
     ~s__maximumPayloadCapacity(X0,X1) | 
     ~s__instrument(s__Carry1,X0)) )),
  inference(resolution,[],[f44,f38])).

tff(f50,plain,(
  ( ! [X2,X0,X1] : 
    (~s__instrument(s__Carry1,X0) | 
     ~s__maximumPayloadCapacity(X0,X1) | 
     ~s__measure(s__MyCar,X2) | 
     ~greaterThan(X2,X1)) )),
  inference(resolution,[],[f47,f37])).

tff(f55,plain,(
  ( ! [X0,X1] : 
    (~s__maximumPayloadCapacity(s__Robot1,X0) | 
     ~s__measure(s__MyCar,X1) | 
     ~greaterThan(X1,X0)) )),
  inference(resolution,[],[f50,f27])).

tff(f60,plain,(
  ( ! [X0] : 
    (~s__measure(s__MyCar,X0) | 
     ~greaterThan(X0,s__MeasureFn__1ReFn(100.0,s__PoundMass))) )),
  inference(resolution,[],[f55,f25])).

tff(f65,plain,(
  ~greaterThan(s__MeasureFn__1ReFn(1.5,s__TonMass),
               s__MeasureFn__1ReFn(100.0,s__PoundMass))),
  inference(resolution,[],[f60,f26])).

tff(f71,plain,(
  ( ! [X1] : (~greaterThan(s__MeasureFn__1ReFn(1.5,X1),
                           s__MeasureFn__1ReFn($quotient(100.0,2000.0),X1))) )),
  inference(resolution,[],[f65,f45])).

tff(f72,plain,(
  ( ! [X1] : (~greaterThan(s__MeasureFn__1ReFn(1.5,X1),
                           s__MeasureFn__1ReFn(0.05,X1))) )),
  inference(evaluation,[],[f71])).

tff(f78,plain,(
  ~$less(0.05,1.5)),
  inference(resolution,[],[f72,f32])).

tff(f95,plain,(
  $false),
  inference(evaluation,[],[f78])).
% SZS output end Proof for Robot-small
% ------------------------------
% Version: Vampire 4.2.2 (commit 6588b35 on 2018-07-19 13:39:17 +0200)
% Termination reason: Refutation

% Memory used [KB]: 9722
% Time elapsed: 0.003 s
% ------------------------------
% ------------------------------
% Success in time 99.856 s

\end{verbatim}

\newpage
\clearpage

\bibliographystyle{apalike}
\bibliography{TFFconversion}
 
\end{document}